\documentclass[10pt,twocolumn,letterpaper]{article}

\usepackage{iccv}
\usepackage{times}
\usepackage{epsfig}
\usepackage{graphicx}
\usepackage{amsmath}
\usepackage{amssymb}
\usepackage{subcaption}

\usepackage[breaklinks=true,bookmarks=false]{hyperref}

\iccvfinalcopy % *** Uncomment this line for the final submission

\def\iccvPaperID{****} % *** Enter the ICCV Paper ID here

\ificcvfinal\fi

\begin{document}

%%%%%%%%% TITLE
\title{Unrealistic Feature Suppression for Generative Adversarial Networks}

\author{Sanghun Kim\\
KyungHee University\\
Department of Computer Science and \\ Engineering
Kyunghee University\\
{\tt\small powerkei@naver.com}
% For a paper whose authors are all at the same institution,
% omit the following lines up until the closing ``}''.
% Additional authors and addresses can be added with ``\and'',
% just like the second author.
% To save space, use either the email address or home page, not both
\and
Seungkyu Lee\\
KyungHee University\\
Department of Computer Science and \\ Engineering
Kyunghee University\\
{\tt\small seungkyu@khu.ac.kr}
}

\maketitle
% Remove page # from the first page of camera-ready.
\ificcvfinal\thispagestyle{empty}\fi

%%%%%%%%% ABSTRACT
\begin{abstract}
Due to the unstable nature of minimax game between generator and discriminator, improving the performance of GANs is a challenging task. Recent studies have shown that selected high-quality samples in training improve the performance of GANs. However, sampling approaches which discard samples show  limitations in some aspects such as the speed of training and optimality of the networks.
In this paper we propose unrealistic feature suppression (UFS) module that keeps high-quality features and suppresses unrealistic features. UFS module keeps the training stability of networks and improves the quality of generated images. We demonstrate the effectiveness of UFS module on various models such as WGAN-GP, SNGAN, and BigGAN. By using UFS module, we achieved better Fréchet inception distance and inception score compared to various baseline models. We also visualize how effectively our UFS module suppresses unrealistic features through class activation maps.
\end{abstract}

%%%%%%%%% BODY TEXT
\section{Introduction}
Generative Adversarial Networks (GANs) \cite{ian} have achieved explosive attention and success in various research areas since it was introduced. GANs consist of two adversarial networks (generator and discriminator) that are trained alternately. Discriminator is trained to distinguish between real and generated fake samples. On the other hand, generator is trained based on the feedback from the discriminator to make realistic fake samples. Thanks to its practical performance of adversarial training strategy, GANs have evolved to various image generation methods such as image to image translation \cite{pix2pix, spade, pix2pixHD}, super resolution \cite{superresolution}, text to image generation \cite{StackGAN}, etc. Improving the performance of GANs is a challenging task due to the unstable nature of minimax game.
Sometimes GANs fall in Nash equilibrium early because of the difficulties in balancing between generator and discriminator training. 

One effort of improving GANs is employing attention blocks that previously have shown improved performance in various classification tasks.
Squeeze and excitation \cite{squeeze} propose a channel attention block and CBAM \cite{cbam} adds a spatial attention block to focus not only on channels but also on spatial features showing that the attention approaches are capable of improving the performance of convolutional neural networks.
In generative models, AttnGAN \cite{attngan} embeds encoded text into networks through attention module so that the networks focus more on the related image details. In image to image translation tasks, SelectionGAN \cite{selectiongan} proposes a method of combining multiple candidate images through multi channel attention selection. DAGAN \cite{dagan} shows instance level translation by specifying the area to focus on in an image. Inspired by non-local neural networks \cite{nonlocal}, SAGAN \cite{sagan} proposes self attention module that extracts feature similarity over entire area of an image.

Another group of methods study how to provide useful gradients in the training of GANs. LOGAN \cite{logan} proposes gradient-based latent optimization scheme for GANs. Latent optimization uses custom latent vector which is optimized by the gradient of networks in training process. It uses $G(z')$ obtained by optimized latent vector $z'$ as fake samples rather than $G(z)$ obtained by random latent $z$ decreasing the influence of random distribution on the networks. 
To this end, LOGAN first forwards the $z$ obtained from random distribution to calculate $D(G(z))$. After obtaining $\nabla z$ in backward process, LOGAN gets optimized latent vector $z'$ ($z' = z + \alpha \nabla z$). Finally $z'$ is re-forwarded to compute $D(G(z'))$ training the networks.
LOGAN shows that the latent optimization not only generate high quality samples but also give better direction to optimal generator.
Since LOGAN performs two forward-backward operations in training, training speed is slow compared to other methods.
Top-k GAN \cite{topk} argues that more realistic samples presented by latent optimization improve network performance. It claims that the success of LOGAN is due to the high quality samples produced by optimized latent $z'$. 
Top-k GAN dismisses unrealistic samples decided by critic(discriminator) from batches and adopt only realistic samples in order to select useful gradients in training process thereby improving the performance of GANs. 
Top-k GAN experimentally shows that high-quality samples in training process lead to improved performance on various networks such as WGAN-GP \cite{gulrajani2017improved}, SNGAN \cite{SNGAN}, SAGAN \cite{sagan} and BigGAN \cite{Biggan}. 
Instance selection \cite{instanceselection} increases training speed by reducing the number of data required for training, rather than improving network itself. 
Original data set is mapped to an embedded space by embedding function $F$, and outliers of data manifold is identified using scoring function $H$. An example of outliers is a sample with large background portion compared to foreground. By excluding such outliers from original data set, training speed of GANs has been dramatically improved.

Even though Top-k GAN \cite{topk} has shown that selected high-quality samples improve the performance of GANs producing useful gradients in the training, it does not mean that entire features of bottom samples present unrealistic characteristics. Earlier iterations of the training procedure may see that both top and bottom samples are all unrealistic. When the training is proceeded creating some realistic fake samples, unrealistic fake samples may not be totally random images including some extent of realistic characteristics. 
Therefore selecting useful gradients in terms of samples may not be fully effective and its performance varies along the quality composition of generated fake samples.
Furthermore, top-k sampling uses only a fraction of overall batch slowing down overall training. To alleviate the defect, annealing scheme is adopted starting with full batch size and gradually reducing it over the course of training.
%Despite this method, top-k sampling still shows slower convergence compared to conventional methods when training with complex datasets such as ImageNet. 
\begin{figure}[!t]
\begin{center}
%\fbox{\rule{0pt}{2in} \rule{0.9\linewidth}{0pt}}
\includegraphics[width=1.0\linewidth]{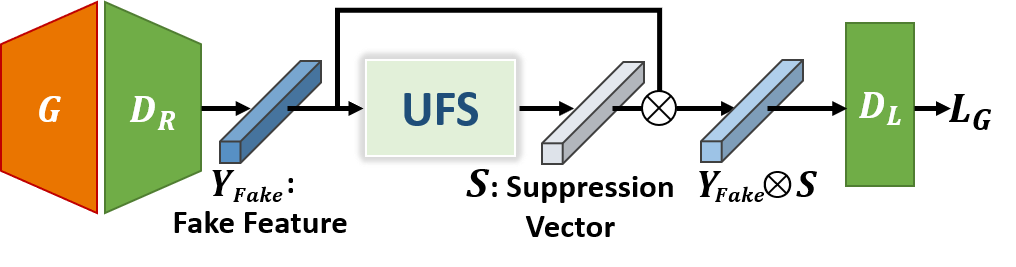}
\end{center}
   \caption{Training Generator with UFS module in GANs: UFS module is applied to generator training process. UFS module derives linear suppression vector $S$ from fake feature $Y_{Fake}$. After multiplying $Y_{Fake}$ by $S$, we forward it to discriminator's last linear layer $D_L$ to calculate $L_G$.}
\label{fig:G_forward}
\end{figure}
Gao \etal \cite{channelpruning} introduces feature boosting and suppression module (FBS) which performs sample unit channel pruning to save computational and memory resources. They argue that different samples have different salient features and dynamic channel pruning is able to amplify salient channels and skip unimportant ones.
Song \etal \cite{finegrained} also introduces a FBSM combined with feature diversification module.

In this work, we propose to perform the selection of useful gradients for improved training of GANs in terms of features that contribute to either realistic or unrealistic samples (or both).  
In order to achieve the goal, we propose unrealistic feature suppression (UFS) module that is able to disregard channels contributing to unrealistic part of fake samples according to the suppression vector shown in figure \ref{fig:G_forward} in generator training step. 
Compared to traditional channel attention blocks that add attention to preferred realistic features, proposed UFS module suppresses selected unrealistic features of all generated fake samples.
Different from feature pruning approaches \cite{channelpruning}, UFS module does not discard selected unrealistic features. Instead it assigns less importance to the features in gradient calculation.
In this way, UFS module keeps the training stability of original networks and improves the quality of generated images. In experimental evaluation, UFS module is embedded on various GANs such as WGAN-GP, SNGAN, BigGAN and tested on various benchmark data sets such as CIFAR-10, CelebA, and ImageNet.

\section{Method}
\begin{figure*}
    \centering
     \begin{subfigure}[b]{0.67\linewidth}
         \centering
         \includegraphics[width=1\linewidth]{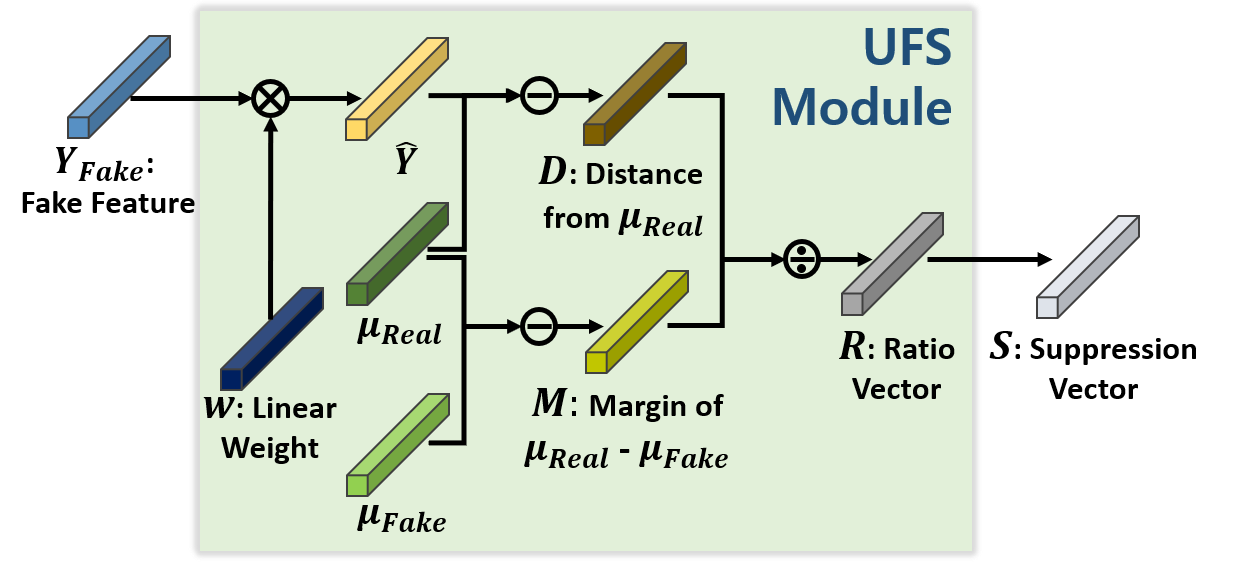}
         \caption{Detailed Structure of UFS module}
         \label{fig: UFSmodule}
     \end{subfigure}
     \hfill
     \begin{subfigure}[b]{0.32\linewidth}
         \centering
         \includegraphics[width=1\linewidth]{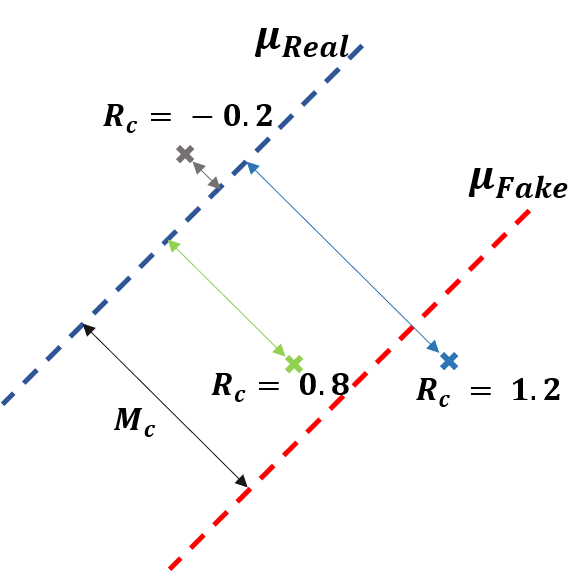}
         \caption{Conceptual Feature Space of a Channel}
         \label{fig: R_M_D}
     \end{subfigure}
        \caption{(a) Unrealistic Feature Suppression (UFS) module: $\mu_{Real}$ and $\mu_{Fake}$ are pre-determined features during training process of discriminator. Using these two feature vectors, we calculate how far $\hat{Y}$ is located from mean of real feature vector $\mu_{Real}$ as the ratio to real-fake margin $M$. (b) Distance ratio $R_c$ and margin $M_c$ of a channel $c$}
        \label{fig:UFSmodule}
\end{figure*}

Figure \ref{fig:G_forward} illustrates how unrealistic feature suppression (UFS) module is incorporated in the training of GANs. Figure \ref{fig:UFSmodule} shows detailed structure of the UFS module with visualized conceptual example.
\subsection{Training Generator with UFS module} \label{sec3.2}
Figure \ref{fig:G_forward} shows a diagram of forward flow using suppression vector $S$ in generator training process. $D$ indicates discriminator. $D_L$ and $D_R$ indicate last and all other preceding layers of the discriminator respectively. We calculate suppression vector $S$ by determining realistic and unrealistic degree in feature level. Fake samples $G(z)$ are mapped to embedded feature $Y_{Fake}$ via $D_R$ ($Y_{Fake} = D_{R}(G(z))$). We forward the feature $Y_{Fake}$ to UFS module to calculate Linear suppression vector $S$. UFS module produces suppression vector $S$ by comparing each feature of $Y_{Fake}$ with corresponding mean of real and fake features. The structure of UFS module is described in detail in Section \ref{sec3.3}. In this way, suppression vector $S$ maintains realistic features of $Y_{Fake}$ and suppresses unrealistic features. 
\begin{equation}
\begin{aligned}
\Bar{y} = D_{L}(Y_{Fake} \otimes UFS(Y_{Fake}))
\end{aligned}
\end{equation}
where $UFS()$ indicates UFS module, and $\otimes$ indicates element wise product. Generator loss $L_G$ is calculated by the expectation of $\Bar{y}$. 
\begin{equation}
\begin{aligned}
L_G = \mathop{\mathbb{E}}_{z{\sim}P_z}[\Bar{y}]
\end{aligned}
\end{equation}

Suppression vector $S$ is enabled in generator training, not in discriminator training.
For discriminator loss $L_D$, we can use both WGAN adversarial loss \cite{wgan} and hinge adversarial loss \cite{geoGAN}.

\subsection{Unrealistic Feature Suppression (UFS) module} \label{sec3.3}
Figure \ref{fig:UFSmodule} shows how linear suppression vector $S$ is extracted from feature vector $Y_{Fake}$. 
In order to distinguish between realistic and unrealistic features, we investigate how each feature of real and fake samples are distributed in the embedded latent space. In general, GANs alternately train discriminator and generator. In the training process of discriminator, real data $x$ and fake data $G(z)$ are forwarded to the latent space obtaining respective feature vectors $Y_{Real}$ and $Y_{Fake}$.
\begin{equation}
\begin{aligned}
Y_{Real} = D_{R}(x),   Y_{Fake} = D_{R}(G(z))
\end{aligned}
\end{equation}
If we consider the last layer of discriminator $D_L$ as a decision function that distinguishes real and fake, we express hyperplane of real and fake as the multiplication between embedded feature and $D_L$. However, what we need is the distributions of fake and real at the feature space, not the sample level. We calculate average feature vectors $\mu_{Real}$ and $\mu_{Fake}$ through element wise product instead of weighted sum calculation. 
\begin{equation}
\begin{aligned}
\mu_{Real} = \frac{1}{n}{\sum_{}^{}}(w \otimes Y_{Real})  ,  \\
\mu_{Fake} = \frac{1}{n}{\sum_{}^{}}(w \otimes Y_{Fake})
\end{aligned}
\end{equation}
where $w$ indicates weight vector of linear layer $D_L$, and $n$ is batch size. UFS module calculates and stores average feature vectors $\mu_{Real}$ and $\mu_{Fake}$ during the discriminator training process. In generator training process, we forward fake sample $G(z)$ to calculate embedded fake feature vector $Y_{Fake}$. $Y_{Fake}$ is used to calculate $\hat{Y}$ through element wise product with $w$. Our objective is to compare each feature, so we don't calculate the average of fake features in training process of generator. 
\begin{equation}
\begin{aligned}
\hat{Y} = w \otimes Y_{Fake}
\end{aligned}
\end{equation}
Now we have criteria $\mu_{Real}$ and $\mu_{Fake}$ to distinguish realistic and unrealistic features. We calculate margin vector $M$ between $\mu_{Real}$ and $\mu_{Fake}$ and distance vector $D$ between $\hat{Y}$ and $\mu_{Real}$ in the feature space. 
\begin{equation}
\begin{aligned}
M = \mu_{Real} - \mu_{Fake}
\end{aligned}
\end{equation}
\begin{equation}
\begin{aligned}
D = \mu_{Real} - \hat{Y}
\end{aligned}
\end{equation}

We obtain distance ratio vector $R$ by dividing distance vector $D$ by margin vector $M$ for each channel separately. $R$ stands for the ratio of how far from the real mean the fake feature is located regarding real-fake margin of each channel separately.
$R_c$ of a channel $c$ is defined as follows.
\begin{equation}
\begin{aligned}
  R_c=\begin{cases}
    \frac{D_c}{M_c} , & \text{if \, $|D_c| \geq \gamma$}.\\
    1, & \text{otherwise}.
  \end{cases}
\end{aligned}
\end{equation}
where $c$ indicates a channel of feature vector. The role of $\gamma$ is to prevent a situation where features that are sufficiently close to real mean being considered unrealistic.
Finally, suppression vector $S$ is created based on $R_c$.
\begin{figure}
    \centering
     \begin{subfigure}[b]{0.49\linewidth}
         \centering
         \includegraphics[width=1\linewidth]{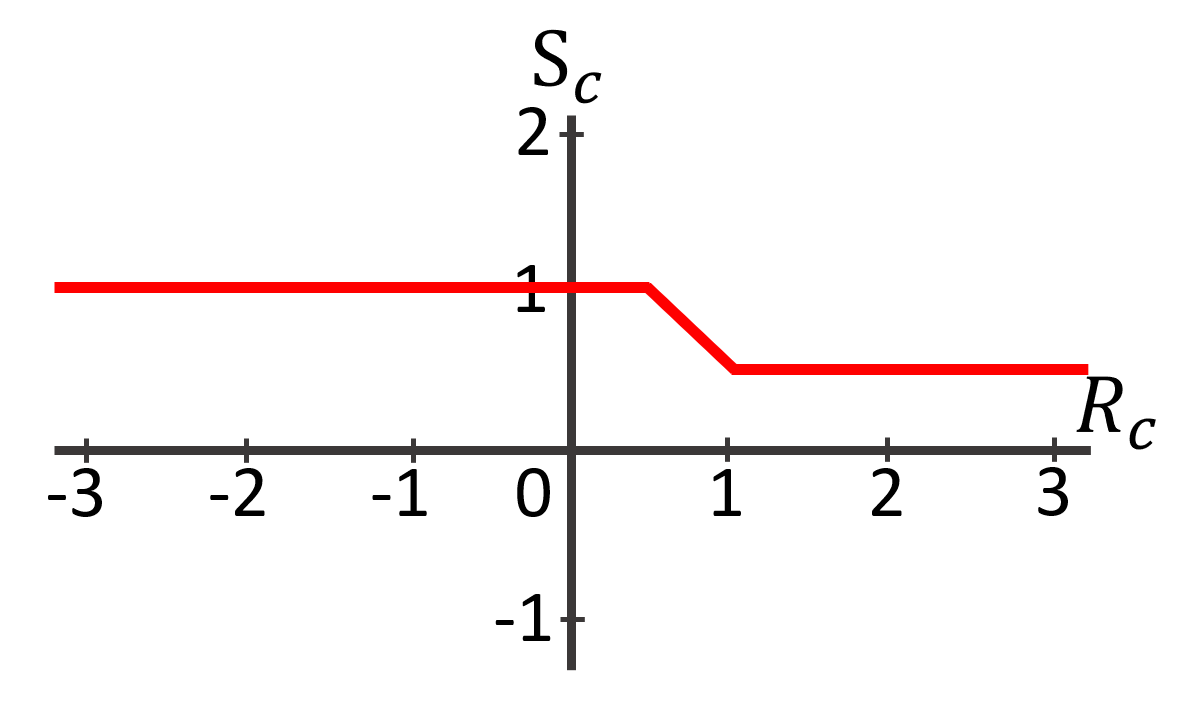}
         \caption{$\alpha=0.5$, $\beta=1$, $\epsilon=1.5$}
         \label{fig: intersection block}
     \end{subfigure}
     \hfill
     \begin{subfigure}[b]{0.49\linewidth}
         \centering
         \includegraphics[width=1\linewidth]{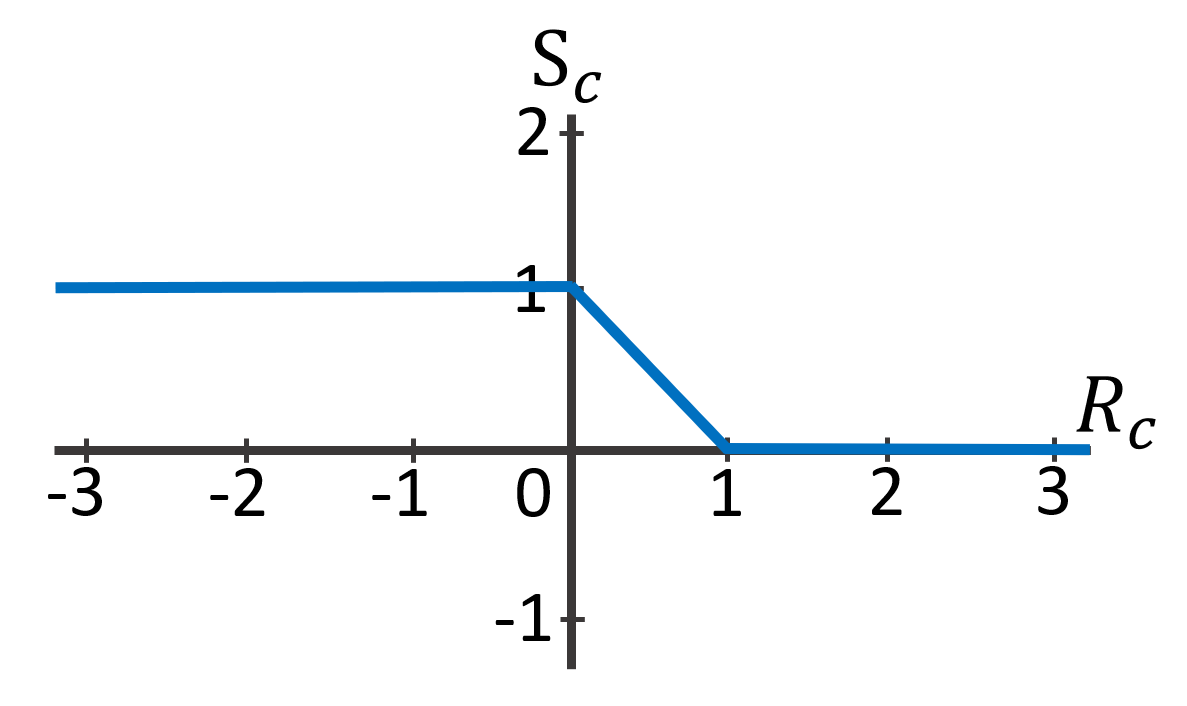}
         \caption{$\alpha=0$, $\beta=1$, $\epsilon=1$}
         \label{fig: intersection block}
     \end{subfigure}
        \caption{Examples of $R_c$ - $S_c$ graph. (a) $\alpha=0.5$, $\beta=1$, $\epsilon=1.5$: Features with $R_c < 0.5$ will be maintained, and features with $R_c >1$ will be suppressed in half. (b) $\alpha=0$, $\beta=1$, $\epsilon=1$: Features with $R_c < 0$ will be maintained, and features with $R_c > 1$ will be dismissed.}
        \label{fig:rcac}
\end{figure}

\begin{equation}
\begin{aligned}
  S_c=\begin{cases}
    -\alpha + \epsilon , & \text{if \, $R_c < \alpha$}.\\
    -R_c + \epsilon, & \text{if \, $\alpha \leq R_c \leq \beta$}. \\
    -\beta + \epsilon, & \text{if \, $R_c > \beta$}.
  \end{cases}
\end{aligned}
\end{equation}
In order to have continuously varying scores in suppression vector according to $R_c$ within a given range, rather than binary values 0 or 1 (use or drop), we define lower and upper bounds $\alpha$ and $\beta$ to constrain $R_c$. $\epsilon$ is added to shift the constrained range to secure appropriate values for suppression vector. Figure \ref{fig:rcac} shows the operating principles of these hyper-parameters.

\section{Experimental Evaluation}
First we run original Top-k GAN \cite{topk} on top-k, bottom-k and random-k samples to compare the performance of the cases and demonstrate the necessity of unrealistic feature suppression rather than realistic sample selection. 
Experimental evaluation of proposed method is performed on CelebA \cite{celeba}, CIFAR10 \cite{cifar}, ImageNet \cite{imagenet}.
We have implemented proposed UFS module on WGAN-GP \cite{gulrajani2017improved}, SNGAN \cite{SNGAN}, BigGAN \cite{Biggan}. 

\subsection{Top-k and Bottom-k Samples}
\label{sampling}
\begin{figure}[!b]
\begin{center}
%\fbox{\rule{0pt}{2in} \rule{0.9\linewidth}{0pt}}
\includegraphics[width=1\linewidth]{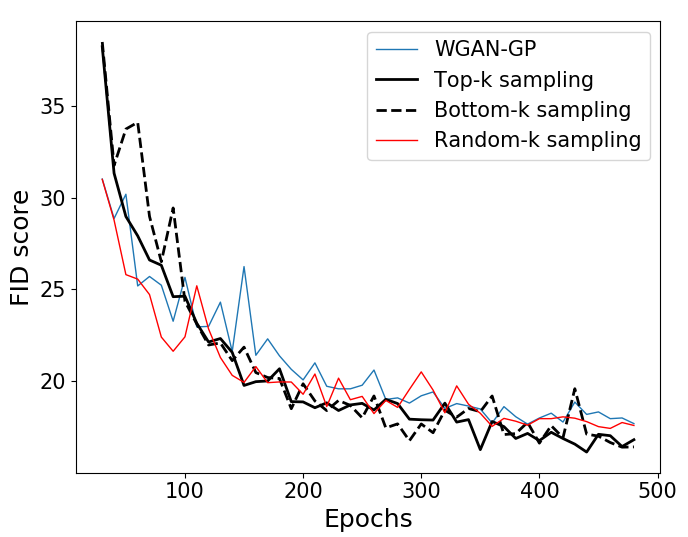}
\end{center}
   \caption{Experiments on various sampling techniques. We train WGAN-GP on CIFAR-10 data set. Unlike other methods showing fluctuation, top-k sampling shows stable convergence. However, the best FID \cite{FID} scores are similar to each other (bottom-k = 15.6 and top-k = 16.1).}
\label{fig:sampling}
\end{figure}
The performance of Top-k GAN \cite{topk} highly depends on a hyper-parameter selecting high-quality samples.
It experimentally has shown with Gaussian mixture model that bottom-K samples produce gradients of negative direction to optimal generator. However, that does not mean that bottom samples always contribute to bad gradients in all process of generator training, especially when it is trained with real image data sets of complex distribution. 
Let's assume that we have a moderately-trained GANs. Generated fake samples are divided into relatively realistic and unrealistic sample groups. Now, we have two strategies: training with only top-k samples (concentrating more on making realistic samples to be more realistic), and training with only bottom-k samples (concentrating more on increasing the quality of unrealistic samples). 
%In this case, unrealistic samples are disclaimed because it is judged that it is impossible to make them more quality due to the limitations of network structure. 
%Second, on the contrary, is to increase the quality of unrealistic samples while maintaining realistic samples as much as possible. In backpropagation process, networks provide gradient to fake samples in proportion to the influence that contribute to the loss. 
%In this aspect, if the latter method is useful for the optimal convergence of the our networks, in other words, top-k sampling may rather interrupt learning if structure of the our networks has the ability to further improve unrealistic samples. 
%For example, if a well-trained networks produces top samples which are close to manifold of real, in other words, if generator produces very realistic samples that are difficult for discriminator to distinguish, top-k sampling only offers very low loss which slow networks training.

We conduct an experiment assuming that bottom samples have ability to provide good gradients. Figure \ref{fig:sampling} shows our experimental results on CIFAR-10 data set. Top-k and bottom-k use only $k$ numbers of top/bottom samples in training process. Batch size is 64 and sampling hyper-parameter $k$ is set to 32. Modified WGAN-GP \cite{gulrajani2017improved} is trained for 500 epochs. FID score is measured every 10 epochs and 10k of real and fake samples are used. Modified WGAN-GP replaces output size of discriminator 1x1 by 8x8. It has the effect of data augmentation that allows last layer of discriminator $D_L$ to see more diverse features leading Modified WGAN-GP to improved performance. In Figure \ref{fig:sampling}, top-k sampling is more stable with less fluctuation than bottom-k sampling, but final FID score is similar to each other.
We also test random-k sampling to see if sampling method itself produces good results regardless of selected top-k or bottom-k samples. Random-k sampling initially seems to converge quickly, but after large fluctuations in the middle it shows similar result to original method. 
In this test, we experimentally observe that bottom-k samples are also capable of providing good gradients for training even though they also bring a risk of negative gradients that are appeared as increased fluctuation in the training.

\subsection{CIFAR-10}
Verification on CIFAR-10 is conducted using both modified WGAN-GP \cite{gulrajani2017improved} and SNGAN \cite{SNGAN}. SNGAN results are discussed in section 4.1.
Modified WGAN-GP that we have used in this test shows improved performance as noted in previous subsection.

\textbf{Experiments on WGAN-GP} Improved WGAN-GP is used as a baseline in this test and table \ref{table:wgancifar} summarizes experimental results.
First, WGAN-GP is combined with Top-k, Bottom-k, and random-k sample selections that are compared to proposed UFS module implemented on the WGAN-GP. In these cases, proposed (WGAN-GP + UFS module) shows best FID score 15.67. We also combine WGAN-GP with UFS module and Top-k, Bottom-k, and random-k sample selection methods where (WGAN-GP + Top-k + UFS module) shows best FID score 15.87. 
These combinations are possible because instance sampling methods (Top-k, Bottom-k, and random-k) select useful gradients in sample level after forwarding step is performed. On the other hand, our UFS module selects useful gradients in latent feature space during forwarding step.
Therefore, combined use of both schemes finds useful gradients in both sample and feature levels simultaneously.  
All experiments are trained for 500 epochs and FID scores are measured using 10k fake and real samples every 10 epochs. Hyper-parameters of UFS module are $\alpha=0$, $\beta=1$, $\epsilon=1$ and $\gamma=0.0001$.
\begin{table}
\begin{center}
\begin{tabular}{|l|c|c|}
\hline
\textbf{CIFAR-10} & FID score\\
\hline\hline
WGAN-GP & 17.56 \\
WGAN-GP + Random-k & 17.29 \\
WGAN-GP + Bottom-k & 15.62 \\
WGAN-GP + Top-k \cite{topk} & 16.13 \\
WGAN-GP + \textbf{UFS (Ours)} & \textbf{15.67} \\
\hline
WGAN-GP + Random-k + \textbf{UFS (Ours)} & 16.10 \\
WGAN-GP + Bottom-k + \textbf{UFS (Ours)} & 16.64 \\
WGAN-GP + Top-k + \textbf{UFS (Ours)} & \textbf{15.87} \\
\hline
\end{tabular}
\end{center}
\caption{FID scores of WGAN-GP trained on CIFAR-10}
\label{table:wgancifar}
\end{table}

\subsection{CelebA}
CelebA data set is also tested with improved WGAN-GP as a baseline.
Networks in all tests are trained for 100 epochs using WGAN adversarial loss. 
Hyper-parameters of UFS module are $\alpha=0$, $\beta=1$, $\epsilon=1$, $\gamma=0.0001$. Batch size is 64 and hyper-paramter $k$ used in top-k GAN is adaptively set starting from 64 and gradually decreasing to 32.
Table \ref{table:wganceleba} compares FID scores. FID scores are calculated every epoch using 10k fake and real samples. (WGAN-GP + UFS module) shows best FID score 6.51 compared to 6.90 of WGAN-GP and 7.96 of Top-k GAN.
\begin{table}
\begin{center}
\begin{tabular}{|l|c|}
\hline
\textbf{CelebA} & FID score\\
\hline\hline
WGAN-GP & 6.90 \\
WGAN-GP + Top-k \cite{topk} & 7.96 \\
WGAN-GP + Top-k + \textbf{UFS (Ours)} & 7.81 \\
WGAN-GP + \textbf{UFS (Ours)} & \textbf{6.51} \\
\hline
\end{tabular}
\end{center}
\caption{FID scores of WGAN-GP trained on CelebA.}
\label{table:wganceleba}
\end{table}

\subsection{ImageNet}
ImageNet is a large-scale data set containing 1.2M images of 1000 classes. Because of the huge size of the data set, training a network with original ImageNet takes long time. For example, training $128 \times 128$ data set takes 2 weeks with 8 NVIDIA V100 GPUs, and $256 \times 256$ requires much longer time. 
ImageNet is also sensitive to batch size. Brock \etal \cite{Biggan} reported that increasing batch size from 256 to 2048 leads performance improvement with FID and IS  \cite{Inceptionscore}. 
Instead of using all of data with large batch size, we follow the method of Miyato \etal \cite{SNGAN} and Devris \etal \cite{instanceselection}. Miyato \etal use a subset of ImageNet named ImageNet dog and cat. Each class of ImageNet dog and cat shows similar characteristics, so networks can be trained easier than using entire ImageNet classes. Devris \etal introduces instance selection which accelerates training speed by reducing the number of samples. We test BigGAN on $64 \times 64$ ImageNet and $128 \times 128$ ImageNet with 50\% instance selection.

\begin{figure}[t]
    \centering
     \begin{subfigure}[b]{0.49\linewidth}
         \centering
         \includegraphics[width=1\linewidth]{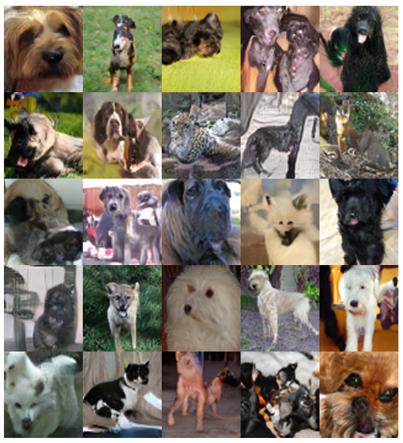}
         \caption{SNGAN + Top-k + \textbf{UFS}}
         \label{fig: intersection block}
     \end{subfigure}
     \hfill
     \begin{subfigure}[b]{0.49\linewidth}
         \centering
         \includegraphics[width=1\linewidth]{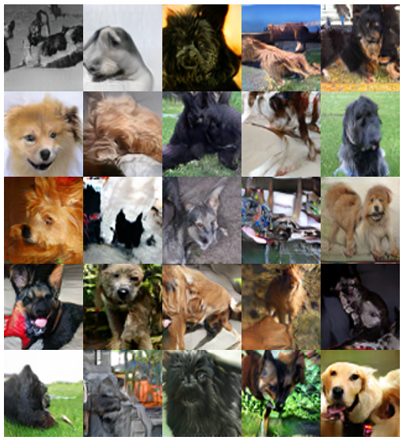}
         \caption{SNGAN + Top-k \cite{topk}}
         \label{fig: intersection block}
     \end{subfigure}
        \caption{Random samples from SNGAN trained on $64 \times 64$ dog and cat images}
        \label{fig:catdog}
\end{figure}

\textbf{ImageNet Dog and cat} \, SNGAN trains their networks by selecting 143 classes out of 1000 classes (total 180k images). All of the selected classes are animal classes corresponding to species of dog and cat, and training parameters are same with SNGAN's baseline. For all experiments we use 64 batch, and hyperparameters of our method are $\alpha=1$, $\beta=1.5$, $\epsilon=2$ and $\gamma=0.0001$. $k$ for Top-k sampling experiments is set to start at 64 and gradually decrease to 32. We train all models for 250k iterations. Table \ref{table:catdog} and figure \ref{fig:catdog} shows our experimental results. Experiment combining Top-k sampling and UFS achieves the best FID score(18.84).
\begin{table}
\begin{center}
\begin{tabular}{|l|c|c|}
\hline
\textbf{ImageNet Dog and Cat} & FID score & IS score \\
\hline\hline
SNGAN \cite{SNGAN} & 20.02 & 11.01 \\
SNGAN + \textbf{UFS (Ours)} & 19.59 & 11.66 \\
SNGAN + Top-k \cite{topk} & 18.99 & 11.86\\
SNGAN + Top-k + \textbf{UFS (Ours)} & \textbf{18.84} & \textbf{12.55} \\
\hline
\end{tabular}
\end{center}
\caption{FID and IS scores on ImageNet dog and cat: Baseline model is SNGAN and all models are trained for 250k iterations.}
\label{table:catdog}
\end{table}

\textbf{ImageNet} {$\mathbf{64 \times 64}$} \, Devris \etal \cite{Biggan} reduces the size of the data set by removing out-liers. Reducing the size of training data set causes small performance degradation, but it shows a dramatic improvement in training speed. Devris \etal reported that instance selection with too large batch size causes performance degradation. 
In this test, we train BigGAN with instance selection using 256 batch instead 2048. $k$ for Top-k sampling experiments is set to start at 256 and gradually decrease to 128. For instance selection, we use inception v3 for data embedding function $F$, and Gaussian model is used for scoring function $H$. Retention ratio is 50\%, \ie we only use 50\% of data for training. All experiments are trained until mode collapse occurred. 

\begin{figure}
\begin{center}
%\fbox{\rule{0pt}{2in} \rule{0.9\linewidth}{0pt}}
\includegraphics[width=0.9\linewidth]{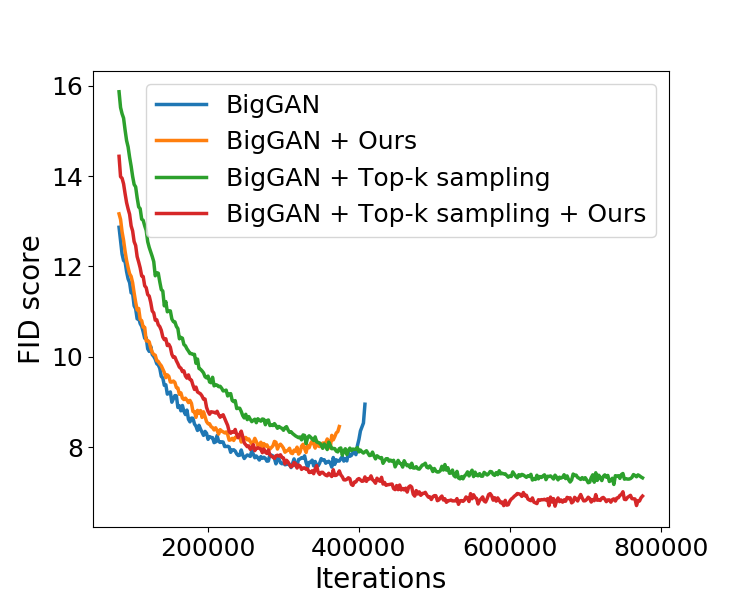}
\end{center}
   \caption{FID scores on $64 \times 64$ ImageNet experiments: BigGAN and (BigGAN + UFS) fall into mode collapse before 400k iterations.}
\label{fig:64graph}
\end{figure}

\begin{figure}[t]
    \centering
     \begin{subfigure}[b]{0.49\linewidth}
         \centering
         \includegraphics[width=1\linewidth]{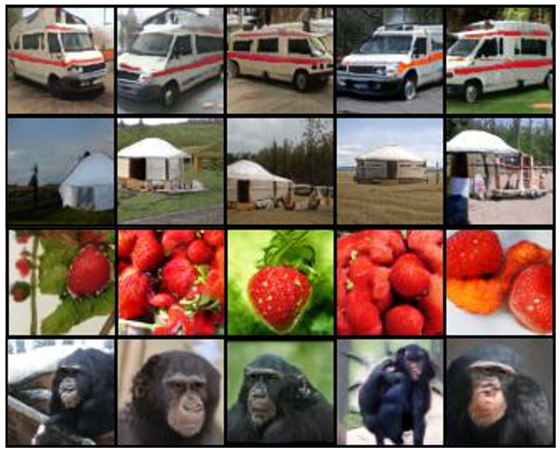}
         \caption{BigGAN + Top-k + \textbf{UFS}}
     \end{subfigure}
     \hfill
     \begin{subfigure}[b]{0.49\linewidth}
         \centering
         \includegraphics[width=1\linewidth]{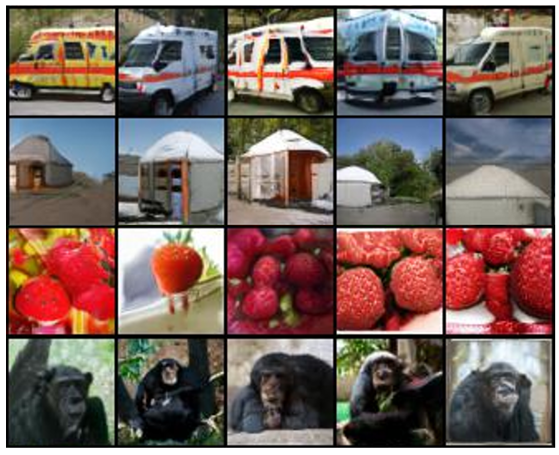}
         \caption{BigGAN + Top-k \cite{topk}}
     \end{subfigure}
        \caption{Samples from BigGAN trained on $64 \times 64$ ImageNet. We pick the classes that clearly show difference in generated image quality.}
        \label{fig:ImageNet64}
\end{figure}

Figure \ref{fig:64graph}, \ref{fig:ImageNet64}, and table \ref{table:ImageNet64} show our results of BigGAN trained on 50\% of ImageNet. 
(BigGAN + UFS module) achieves 7.84 FID score that is worse than original BigGAN(7.58). Top-k sampling achieves better FID score(7.17), but it shows very slow training speed compared to original method. While original method takes 310k iterations to reach highest FID 7.58, Top-k takes 460k iterations. On the other hand, (BigGAN + Top-k + UFS module) shows much faster speed than only using top-k sampling. As you can see in Figure \ref{fig:64graph}, initial training speed is slower than the original, but faster than Top-k sampling. To reach the FID score 10, the original takes 128k iterations, (BigGAN + Top-k + UFS module) takes 154k, and Top-k takes 194k. When comparing the best FID score, (BigGAN + Top-k + UFS module) achieves a much better FID score(6.73) than other methods. It is noteworthy that experiments without Top-k sampling fall into mode collapse, while experiments using Top-k sampling don't show mode collapse even if they are trained more than 2 times longer iterations.

Since FID score is highly sensitive to variance of model distribution, we need to check other metrics for evaluating our model. Precision \& Recall \cite{improvedPrecisionRecall, PrecisionRecall} and Density \& Coverage \cite{DensityCoverage} check the distribution manifold through k-NN on embedded space. We choose the best model based on FID score of (BigGAN + Top-k) and (BigGAN + Top-k + UFS module). Except density metric, (BigGAN + Top-k + UFS module) achieves better scores than Top-k: precision(\textbf{0.8547} vs 0.8456), recall(\textbf{0.6086} vs 0.5960), density(1.303 vs \textbf{1.306}) and coverage(\textbf{0.9406} vs 0.9325). (BigGAN + Top-k + UFS module) achieves better precision, but density is lower than (BigGAN + Top-k), so it is hard to say that which one is better in terms of the overlap ratio in real data manifold of generated samples. But (BigGAN + Top-k + UFS module) achieves better recall and coverage, which represents (BigGAN + Top-k + UFS module) generates more diverse images which overlap real data manifold.

\begin{table}
\begin{center}
\begin{tabular}{|l|c|c|}
\hline
\textbf{ImageNet 64 $\mathbf{\times}$ 64} & FID score & IS score \\
\hline\hline
BigGAN \cite{Biggan} & 7.58 & 49.24 \\
BigGAN + \textbf{UFS (Ours)} & 7.84 & 47.17 \\
BigGAN + Top-k \cite{topk} & 7.17 & 45.05\\
BigGAN + Top-k + \textbf{UFS (Ours)} & \textbf{6.73} & \textbf{49.93} \\
\hline
\end{tabular}
\end{center}
\caption{FID and IS scores of BigGAN trained on $64 \times 64$ ImageNet: To accelerate the speed of all experiments, we use instance selection to reduce the number of training data by 50\%.}
\label{table:ImageNet64}
\end{table}

\begin{figure}[t]
    \centering
     \begin{subfigure}[b]{0.49\linewidth}
         \centering
         \includegraphics[width=1\linewidth]{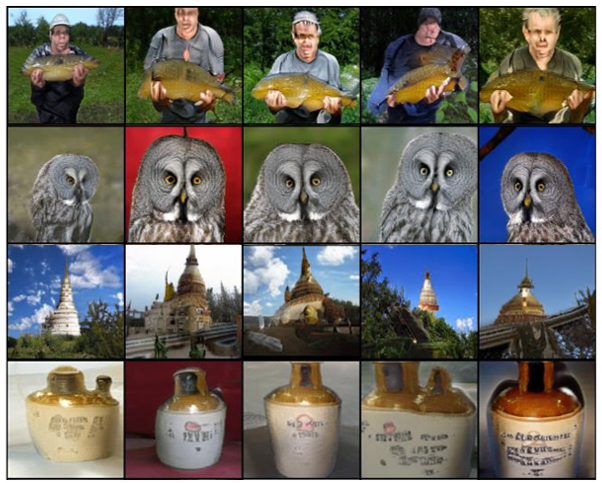}
         \caption{BigGAN + Top-k + \textbf{UFS}}
     \end{subfigure}
     \hfill
     \begin{subfigure}[b]{0.49\linewidth}
         \centering
         \includegraphics[width=1\linewidth]{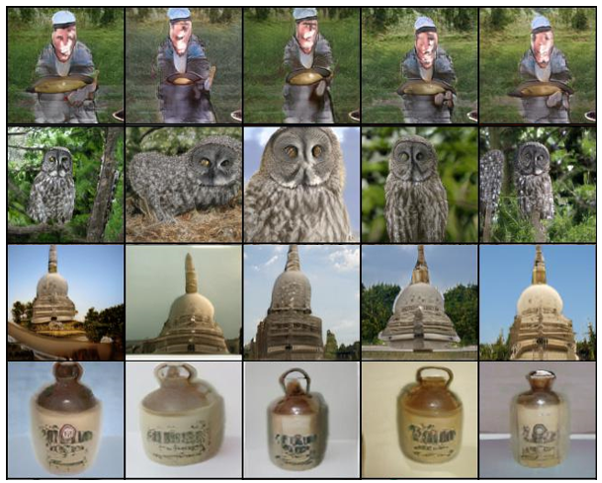}
         \caption{BigGAN + Top-k \cite{Biggan}}
     \end{subfigure}
        \caption{Samples from BigGAN trained on $128 \times 128$ ImageNet. We pick the classes that clearly show difference in generated image quality.}
        \label{fig:ImageNet128}
\end{figure}

\textbf{ImageNet} {$\mathbf{128 \times 128}$} \, To check the performance of our method with higher resolution images, we train BigGAN on $128 \times 128$ ImageNet. We use instance selection to speed up training. Hyper-parameters for instance selection are same to experiments on ImageNet $64 \times 64$. Batch size is 128, but we accumulate gradient twice following the experiments of Devris \etal \cite{instanceselection}. $k$ is set to start at 128 and gradually decrease to 64. To accelerate training speed, we also anneal $\beta$ gradually from 1 to 1.5, it means that we use full feature in early training and gradually suppress unrealistic features over epochs. Other hyper-parameters are the same as previous experiments, $\alpha=1$, $\epsilon=2$, $\gamma=0.0001$.
We train all models for 800k iterations. Figure \ref{fig:ImageNet128} and table \ref{table:ImageNet128} show test results of BigGAN trained on 50\% of ImageNet $128 \times 128$. Similar to $64 \times 64$ experiments, Top-k sampling shows very slow training speed, which results in insufficient convergence for 800k iterations. BigGAN and (BigGAN + UFS module) fall into mode collapse before 400k iterations. (BigGAN + Top-k + UFS module) shows best FID score 8.78 and inception score 111.01. 
We also check other metrics for evaluating our models. (BigGAN + Top-k + UFS module) achieves better scores than Top-k sampling: precision(\textbf{0.8963} vs 0.8852), recall(\textbf{0.5488} vs 0.5395), density(\textbf{1.4594} vs 1.3917), and coverage(\textbf{0.9297} vs 0.9101).
\begin{table}
\begin{center}
\begin{tabular}{|l|c|c|}
\hline
\textbf{ImageNet 128 $\mathbf{\times}$ 128} & FID score & IS score \\
\hline\hline
BigGAN \cite{Biggan} & 9.88 & 108.40 \\
BigGAN + \textbf{UFS (Ours)} & 11.08 & 94.83 \\
BigGAN + Top-k \cite{topk} & 9.76 & 108.86\\
BigGAN + Top-k + \textbf{UFS (Ours)} & \textbf{8.78} & \textbf{111.01} \\
\hline
\end{tabular}
\end{center}
\caption{FID and IS scores of BigGAN trained on $128 \times 128$ ImageNet: To accelerate the speed of all experiments, we use instance selection to reduce the number of data by 50\%.}
\label{table:ImageNet128}
\end{table}

\subsection{How does suppression vector work?}
To intuitively understand how $S$ helps training, we need to visualize what discriminator actually see. Class activation map \cite{CAM} visualizes which regions of the image that CNN sees and makes judgments. We obtain class activation map in three different ways from pretrained BigGAN networks. We calculate class activation maps based on the equations below. 
\begin{figure*}
\begin{center}
%\fbox{\rule{0pt}{2in} \rule{0.9\linewidth}{0pt}}
\includegraphics[width=0.95\linewidth]{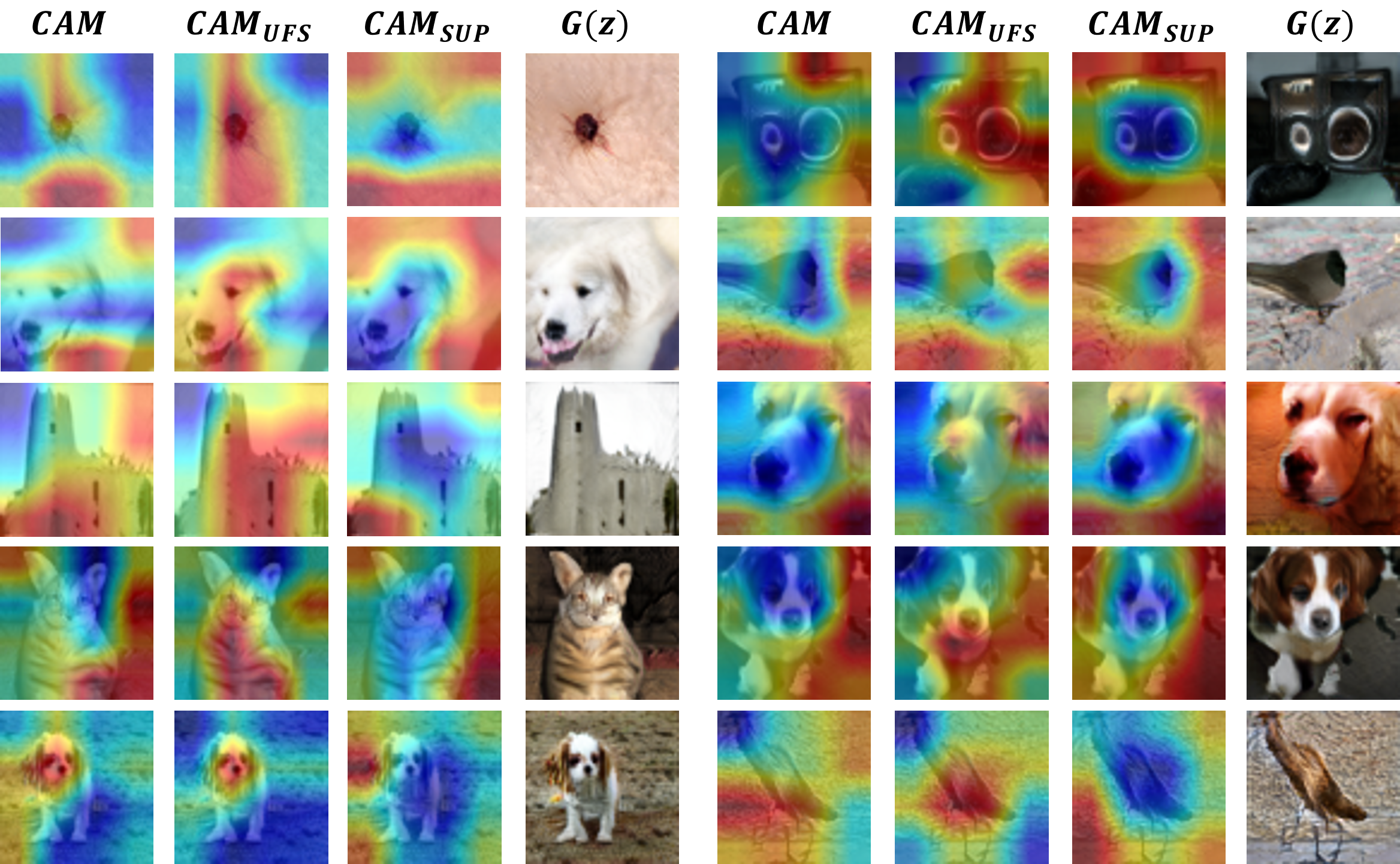}
\end{center}
   \caption{Visualization on how suppression vector $S$ works on discriminator. $CAM$ is general class activation map of discriminator, which shows most realistic area in images. $CAM_{UFS}$ and $CAM_{SUP}$ are class activation maps of discriminators with suppression vector $S$ and suppressed vector $1 - S$, respectively.}
\label{fig:cam}
\end{figure*}
\begin{equation}
\begin{aligned}
  CAM^{i,j} &= \left\langle \Tilde{Y}_{i,j}, w \right\rangle  \\
  CAM_{UFS}^{i,j} &= \left\langle \Tilde{Y}_{i,j} \otimes S, w \right\rangle  \\
  CAM_{SUP}^{i,j} =& \left\langle \Tilde{Y}_{i,j} \otimes (1 - S), w \right\rangle 
\end{aligned}
\end{equation}
where $\tilde{Y}$ represents feature before global sum pooling is applied. Since WGAN based adversarial loss leads $D(x)$ to be greater than $D(G(z))$, class activation map represents the most realistic areas in terms of discriminator. $CAM_{SUP}$ shows the regions of the most unrealistic features in terms of our UFS module. Therefore, $CAM_{SUP}$ shows regions that will be suppressed by our methods. On the other hand, $CAM_{UFS}$ shows class activation map after unrealistic features are suppressed. Therefore, $CAM_{UFS}$ shows the regions of realistic features in terms of UFS module.
%You can see that the result of subtracting activated part of $CAM_{SUP}$ from activated part of $CAM$ is $CAM_{UFS}$. 

Figure \ref{fig:cam} shows sample activation maps. 
$CAM$ includes foreground with unique and outstanding appearance as well as background with relatively rough and ambiguous visual patterns. Since the appearance of such background is easier to learn, $CAM$ tends to include unrealistic background regions that are relatively easy to be learned. 
However, as we observe the $CAM_{UFS}$ examples in figure \ref{fig:cam}, UFS module suppresses features that are far from average real feature $\mu_{Real}$, actually suppressing unrealistic background region and promoting discriminator to concentrate more on realistic foreground.
Similar to Top-k GAN which ignores unrealistic samples and trains networks to make realistic samples better, UFS module ignores unrealistic features and trains networks to make realistic features better.

\section{Discussion and Conclusion}

%\subsection{SNGAN+UFS module on CIFAR-10} 
\subsection{Why unrealistic feature suppression instead of dismission?}
\begin{figure}
    \centering
     \begin{subfigure}[b]{0.49\linewidth}
         \centering
         \includegraphics[width=1\linewidth]{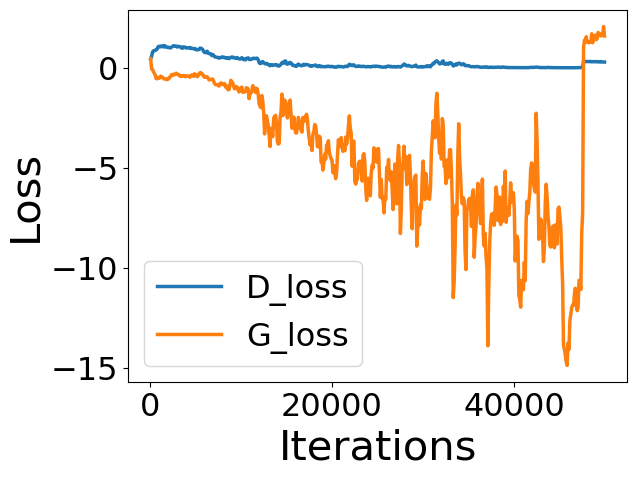}
         \caption{$\alpha=0$, $\beta=1$, $\epsilon=1$}
         \label{fig: intersection block}
     \end{subfigure}
     \hfill
     \begin{subfigure}[b]{0.49\linewidth}
         \centering
         \includegraphics[width=1\linewidth]{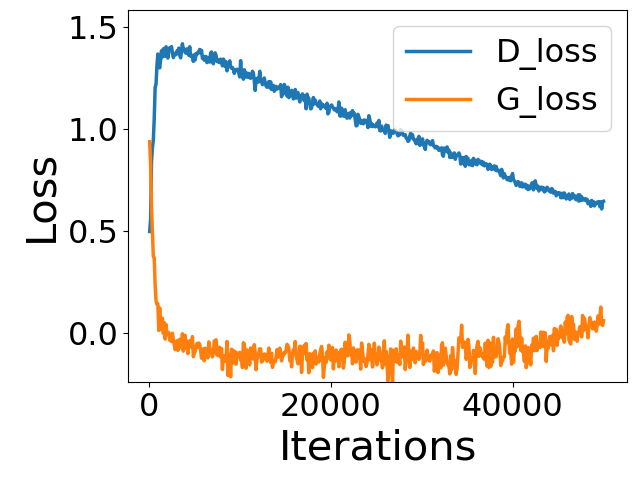}
         \caption{$\alpha=1$, $\beta=1.5$, $\epsilon=2$}
         \label{fig: intersection block}
     \end{subfigure}
        \caption{Training loss of SNGAN experiments on CIFAR-10. (a) $\alpha=0$, $\beta=1$ and $\epsilon=1$: Discriminator converges in early iterations, and after that generator loss becomes unstable and networks go to mode collapse. (b) $\alpha=1$, $\beta=1.5$ and $\epsilon=2$: It constrains $S_c$ in space of $[0.5, 1]$.}
        \label{fig:SNGAN_loss}
\end{figure}

SNGAN + UFS module tested on CIFAR-10 shows unstable training. Figure \ref{fig:SNGAN_loss} shows generator loss and discriminator loss of entire training process. While the discriminator is quickly trained, the generator loses its direction for training. We assume that when a discriminator becomes too powerful, \ie, a discriminator easily distinguishes between real and fake features, UFS module looks dismiss most of unrealistic features. We adjust hyper-parameters to suppress unrealistic features rather than dismissing them. By setting $0 < \epsilon - \beta < 1$, unrealistic features are suppressed. Table \ref{table:hyper} shows the summary of the our experiments. We observed that dismissing unrealistic features gives rise to severe mode collapse. On contrary, suppressing unrealistic features doesn't fall into mode collapse. The optimal hyper-parameters empirically found are $\alpha=1$, $\beta=1.5$ and $\epsilon=2$.

%\begin{figure}
%\begin{center}
%\fbox{\rule{0pt}{2in} \rule{0.9\linewidth}{0pt}}
%\includegraphics[width=1.0\linewidth]{LaTeX/fig6.png}
%\end{center}
%   \caption{Training loss of SNGAN experiments on CIFAR-10. (a) shows the case set to $\alpha=0$, $\beta=1$ and $\epsilon=1$. Discriminator converges in early iterations, and after that generator loss falters unstably and networks goes to mode collapse. This means that our method drop unrealistic features too strong so that generator can't get proper gradients for training. (b) shows the case set to $\alpha=1$, $\beta=1.5$ and $\epsilon=2$ which constrains $S_c$ in space of $[0.5, 1]$. We can see (b) can train more stably compared to previous.}
%\label{fig:SNGAN_loss}
%\end{figure}

\begin{table}[!b]
\begin{center}
\begin{tabular}{|c|c|c|c|c|c|}
\hline
$\alpha$ & $\beta$ & $\epsilon$ & FID score & IS score & UFS \\
\hline\hline
0 & 1 & 1 & 43.37 & 6.20 & Dismission\\
1 & 2 & 2.5 & 19.47 & 7.93 & \textbf{Suppression}\\
1 & 2 & 3 & 19.54 & 8.26 & \textbf{Suppression}\\
1 & 3 & 3 & 50.57 & 6.70 & Dismission\\
\hline
1 & 1.2 & 2 & 18.07 & 8.09 & \textbf{Suppression}\\
1 & 1.3 & 2 & 19.29 & 7.92 & \textbf{Suppression}\\
1 & 1.4 & 2 & 18.82 & 8.09 & \textbf{Suppression}\\
1 & 1.5 & 2 & \textbf{17.10} & \textbf{8.50} & \textbf{Suppression}\\
\hline
\end{tabular}
\end{center}
\caption{Experiments to get proper hyperparameters for our method. Baseline model is SNGAN, and dataset is CIFAR-10. We train all models with 250k iterations. Suppression means suppressing unrealistic features rather than dismissing them by setting ($0 < \epsilon - \beta < 1$).}
\label{table:hyper}
\end{table}

\subsection{Conclusion}
In this work, we have proposed unrealistic feature suppression (UFS) module that suppress unrealistic features in generator training. Effectiveness of UFS module has been proved through extensive experimental evaluations on various backbone networks such as WGAN-GP, SNGAN, BigGAN. In ImageNet experiments, we show that a method combining Top-k selection and UFS module converges faster and better compared to prior methods. 
{\small
\bibliographystyle{ieee_fullname}
\bibliography{egbib}
}

\end{document}